\documentclass[10pt,twocolumn,letterpaper]{article}

\usepackage[pagenumbers]{iccv} 

\usepackage{graphicx}
\usepackage{amsmath}
\usepackage{amssymb}
\usepackage{booktabs}
\usepackage[accsupp]{axessibility} 
\usepackage{multirow, makecell}
\usepackage{float}
\usepackage{xcolor}
\usepackage{array}

\usepackage{comment}
\usepackage{amsthm}

\definecolor{iccvblue}{rgb}{0.21,0.49,0.74}
\usepackage[pagebackref,breaklinks,colorlinks,allcolors=iccvblue]{hyperref}

\def\paperID{8178} 
\def\confName{ICCV}
\def\confYear{2025}

\title{D$^2$ST-Adapter: Disentangled-and-Deformable Spatio-Temporal \\ Adapter for Few-shot Action Recognition}

\author{
Wenjie Pei$^{*\dagger\,1,2}$\hspace{0.5cm}
Qizhong Tan$^{*\,1}$\hspace{0.5cm}
Guangming Lu$^{1}$\hspace{0.5cm}
Jiandong Tian$^{3}$\hspace{0.5cm}
Jun Yu$^{1}$\\
$^1$Harbin Institute of Technology, Shenzhen\\
$^2$Peng Cheng Laboratory\\
$^3$Shenyang Institute of Automation, Chinese Academy of Sciences\\
{\tt\small wenjiecoder@outlook.com, 24B951007@stu.hit.edu.cn}
}

\begin{document}
\maketitle
\let\thefootnote\relax\footnotetext{$*$ Equal contribution. \hspace{1mm} $\dagger$ Corresponding author.}
\begin{abstract}
Adapting pre-trained image models to video modality has proven to be an effective strategy for robust few-shot action recognition. 
In this work, we explore the potential of adapter tuning in image-to-video model adaptation and propose a novel video adapter tuning framework, called Disentangled-and-Deformable Spatio-Temporal Adapter (D$^2$ST-Adapter). It features a lightweight design, low adaptation overhead and powerful spatio-temporal feature adaptation capabilities. D$^2$ST-Adapter is structured with an internal dual-pathway architecture that enables built-in disentangled encoding of spatial and temporal features within the adapter, seamlessly integrating into the single-stream feature learning framework of pre-trained image models.
In particular, we develop an efficient yet effective implementation of the D$^2$ST-Adapter, incorporating the specially devised anisotropic Deformable Spatio-Temporal Attention as its pivotal operation. This mechanism can be individually tailored for two pathways with anisotropic sampling densities along the spatial and temporal domains in 3D spatio-temporal space, enabling disentangled encoding of spatial and temporal features while maintaining a lightweight design.
Extensive experiments by instantiating our method on both pre-trained ResNet and ViT demonstrate the superiority of our method over state-of-the-art methods. Our method is particularly well-suited to challenging scenarios where temporal dynamics are critical for action recognition.
Code is available at \url{https://github.com/qizhongtan/D2ST-Adapter}.


\end{abstract}
\begin{figure*}[t]
  \centering
   \includegraphics[width=\linewidth]{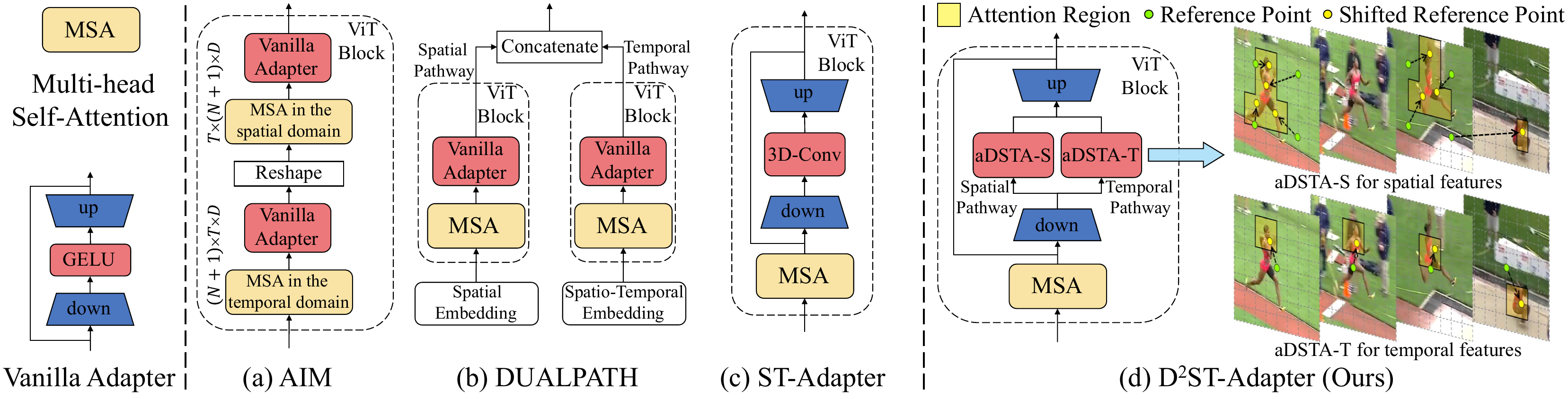}
   \caption{Comparison with previous adapter-based methods. Both AIM and DUALPATH pre-learn the spatial and temporal features separately by duplicating the whole MSA modules, and insert vanilla adapter into each duplicated module for feature adaptation, which incurs heavy-weight model design. ST-Adapter designs the adapter for video data using 3D convolution whose performance is limited by the joint learning of spatio-temporal features. In contrast, our \emph{D$^2$ST-Adapter} is designed in a dual-pathway architecture to adapt the spatial and temporal features in a disentangled manner while maintaining the single-stream feature learning framework of the pre-trained model. Furthermore, we design the anisotropic Deformable Spatio-Temporal Attention (aDSTA), which configures anisotropic sampling densities in the spatial and temporal domains to model two pathways specifically and enables \emph{D$^2$ST-Adapter} to adapt features in a global view.
   }
   \label{fig:introduction}
\end{figure*}

\section{Introduction}
Few-shot action recognition aims to learn an action recognition model from a set of base classes of video data, which can recognize novel categories of actions using only a few support samples. To this end, learning an effective feature extractor that is generalizable across different classes is crucial. A typical way~\cite{OTAM,ARN,TARN,HyRSM,CLIP-FSAR} is to leverage pre-trained large image models~\cite{Resnet,C3D,CLIP} for feature learning by image-to-video model adaptation.

Most existing methods~\cite{TRX,MoLo,CLIP-FSAR} seek to adapt large vision models to few-shot action recognition by fine-tuning either the entire or partial model. While such methods have achieved promising performance, there are two important limitations that hamper them from exploiting the full potential of the pre-trained models. First, fully fine-tuning pre-trained models tends to overfit in few-shot learning scenarios, whereas partial fine-tuning is less effective due to its limited capacity for adaptation. Second, most large vision models are trained on image data due to the scarcity of video data. 
Thus, these fine-tuning based methods require to learn the temporal features as a post-processing step after feature extraction by the pre-trained backbone, typically by constructing an auxiliary module to model the temporal dynamics~\cite{STRM,MTFAN,HyRSM,MoLo}, which has limited effectiveness compared to methods that incorporate temporal learning throughout the entire feature learning stage.

As a prominent parameter-efficient fine-tuning technique~\cite{Adapter,Coop,VPT,he2021towards}, 
adapter tuning~\cite{Adapter,adaptformer,K-Adapter} is particularly well-suited to few-shot learning scenarios due to much lower parameter-learning overhead. 
A straightforward way to apply adapter tuning to image-to-video model adaptation is to employ the vanilla image adapter~\cite{Adapter} to adapt the pre-learned spatial and temporal features separately, which is exactly adopted by AIM~\cite{AIM} and DUALPATH~\cite{DUALPATH}. As shown in Figure~\ref{fig:introduction},  
both AIM and DUALPATH first pre-learn the spatial and temporal features separately by duplicating the multi-head self-attention modules, either in cascaded manner (AIM) or parallel manner (DUALPATH). Subsequently, they insert individual vanilla adapters in each duplicated module for specific feature adaptation. However, such methods inevitably result in a  heavyweight model architecture and substantial computational overhead. As illustrated in Figure~\ref{fig:introduction}, ST-Adapter~\cite{ST-Adapter} is the first to design a specialized adapter for video data by incorporating a 3D convolution layer into the vanilla adapter to learn spatio-temporal features. 
Nevertheless, we investigate two potential limitations of ST-Adapter. 
First, it learns the spatial and temporal features jointly by 3D convolution, whereas it has been shown that in low-data scenarios, such a joint learning paradigm for spatio-temporal features in video data is inferior to learning the spatial and temporal features in a disentangled manner, such as SlowFast~\cite{Slowfast} and other two-stream designs~\cite{Two-stream,TSN}. Second, the convolutional operation, especially with shallow layers for lightweight design, has a limited local receptive field in both the spatial and temporal domains, which also limits the performance of ST-Adapter.

In this work we present the Disentangled-and-Deformable Spatio-Temporal Adapter (\emph{D$^2$ST-Adapter}) 
to tackle the aforementioned limitations. Specifically, we design our \emph{D$^2$ST-Adapter} as a dual-pathway architecture, as illustrated in Figure~\ref{fig:introduction}, in which the spatial pathway is responsible for capturing the appearance features while the temporal pathway focuses on learning the temporal dynamics. This built-in disentangled encoding of spatio-temporal features equips our \emph{D$^2$ST-Adapter} with superior image-to-video model adaptation capabilities while maintaining a lightweight design and low adaptation overhead. As a result, it benefits from a more efficient model design compared to AIM and DUALPATH, which perform separate feature pre-learning and individual adaptation for spatial and temporal features by duplicating learning modules. Meanwhile, our \emph{D$^2$ST-Adapter} demonstrates superior feature learning capability for video data compared to ST-Adapter, owing to its disentangled adaptation scheme.

We develop an efficient yet effective implementation of the D$^2$ST-Adapter, featuring the specially devised anisotropic Deformable Spatio-Temporal Attention (aDSTA) as the pivotal operation to enable built-in disentangled encoding capability. 
It adapts the deformable attention~\cite{DAT} from 2D image space to 3D spatio-temporal space, 
which allows our \emph{D$^2$ST-Adapter} to encode features in both spatial and temporal domains in a global view while keeping lightweight modeling. One novel design of our aDSTA is that it can be tailored with anisotropic sampling densities along spatial and temporal domains, enabling specialized versions of aDSTA to model the spatial and temporal pathways separately. As shown in Figure~\ref{fig:introduction}, we tailor aDSTA-T with denser sampling along temporal domain than spatial domain for temporal pathway since it focuses on capturing temporal features. In contrast, the tailored aDSTA-S for spatial pathway samples denser points along spatial domain. 

To conclude, we make the following contributions:
\begin{itemize}[leftmargin =*, itemsep = 0pt, topsep = -2pt]
    \item We propose \emph{D$^2$ST-Adapter}, a novel video adapter tuning framework for few-shot action recognition. It is designed in a dual-pathway architecture featuring built-in disentangled adaptation for spatial and temporal features, enabling effective yet efficient image-to-video model adaptation for few shot action recognition.
    \item We devise aDSTA, which can be tailored with anisotropic sampling densities along spatial and temporal domains to model the spatial and temporal pathways separately, yielding a lightweight implementation of \emph{D$^2$ST-Adapter}.
    \item Extensive experiments on five benchmarks with instantiations of our method on pre-trained ResNet~\cite{Resnet} and ViT~\cite{CLIP}, demonstrate the superiority of our method over other methods, particularly in challenging scenarios like SSv2 benchmark where the temporal dynamics are critical for action recognition. 
\end{itemize}
\section{Related Work}

\noindent\textbf{Few-shot Action Recognition.}
%
Most existing few-shot action recognition methods adopt the metric-based paradigm to classify videos, and primarily focus on two directions to deal with this task.
The first one is to investigate the spatio-temporal modeling~\cite{CMN-J,ARN,STRM,MoLo,GgHM,RFPL,AMFAR,kumar2024trajectory}.
%
%
STRM~\cite{STRM} enriches local patch features and global frame features for joint spatio-temporal modeling.
MoLo~\cite{MoLo} designs a motion autodecoder to explicitly extract motion dynamics in a unified network.
Our method also aims to model spatio-temporal features effectively and efficiently based on adapter-tuning technique.
Another direction focuses on designing effective metric learning strategies~\cite{OTAM,TRX,HyRSM,MTFAN,huang,nguyen}.
OTAM~\cite{OTAM} utilizes the DTW~\cite{DTW} algorithm to calculate video distances with strict temporal alignment.
TRX~\cite{TRX} exhaustively enumerates all sub-sequences of support and query videos and matches them using attention mechanism.
HyRSM~\cite{HyRSM} applies a novel bidirectional mean Hausdorff metric (referred to as Bi-MHM) to alleviate the strictly ordered constraints.
We evaluate our \emph{D$^2$ST-Adapter} using the above three classical matching metrics to demonstrate its effectiveness and robustness.

\noindent\textbf{Adapter Tuning.}
As a classical parameter-efficient fine-tuning method, adapter tuning is first proposed in~\cite{Adapter}, and quickly draws attention in many other research areas~\cite{adaptformer,K-Adapter,he2021towards}.
A common practice is to build up a lightweight module (named \emph{Adapter}) which only consisting of negligible learnable parameters, and selectively plug it into a pre-trained model.
During training, only the parameters of inserted adapters are updated while the original model remains frozen, leading to efficient task adaptation.
Recently, some works apply this method to adapt image models for action recognition.
AIM~\cite{AIM} duplicates multi-head self-attention modules and plugs adapters after them to separately learn the spatial and temporal features in a cascaded manner.
Similarly, DUALPATH~\cite{DUALPATH} explicitly builds two-stream architecture upon ViT and utilizes adapters to parallelly learn spatial and temporal features.
ST-Adapter~\cite{ST-Adapter} employs depth-wise 3D convolution to construct adapter for feature adaptation, which endows it with spatio-temporal modeling capability.
Our \emph{D$^2$ST-Adapter} follows the basic framework of ST-Adapter, and meanwhile optimize the essential technical designs for few-shot action recognition. 
\section{Method}
\subsection{Overview}
\noindent\textbf{Problem Formulation of Few-shot Action Recognition.} 
The task of few-shot action recognition typically follows episodic paradigm~\cite{MatchNet,OTAM,HyRSM,MoLo}. 
An episode includes a support set $\mathcal{S}$ consisting of $N$ classes and $K$ labeled samples for each class (referred to as the $N$-way $K$-shot task), as well as a query set $\mathcal{Q}$ that contains unlabeled samples to be classified.
In each episode, we aim to classify every query into one of the $N$ classes with the guidance of the support set. Such episodic task setting is consistently adopted during all the training, validation, and test stages. 

\begin{figure*}[t]
  \centering
   \includegraphics[width=\linewidth]{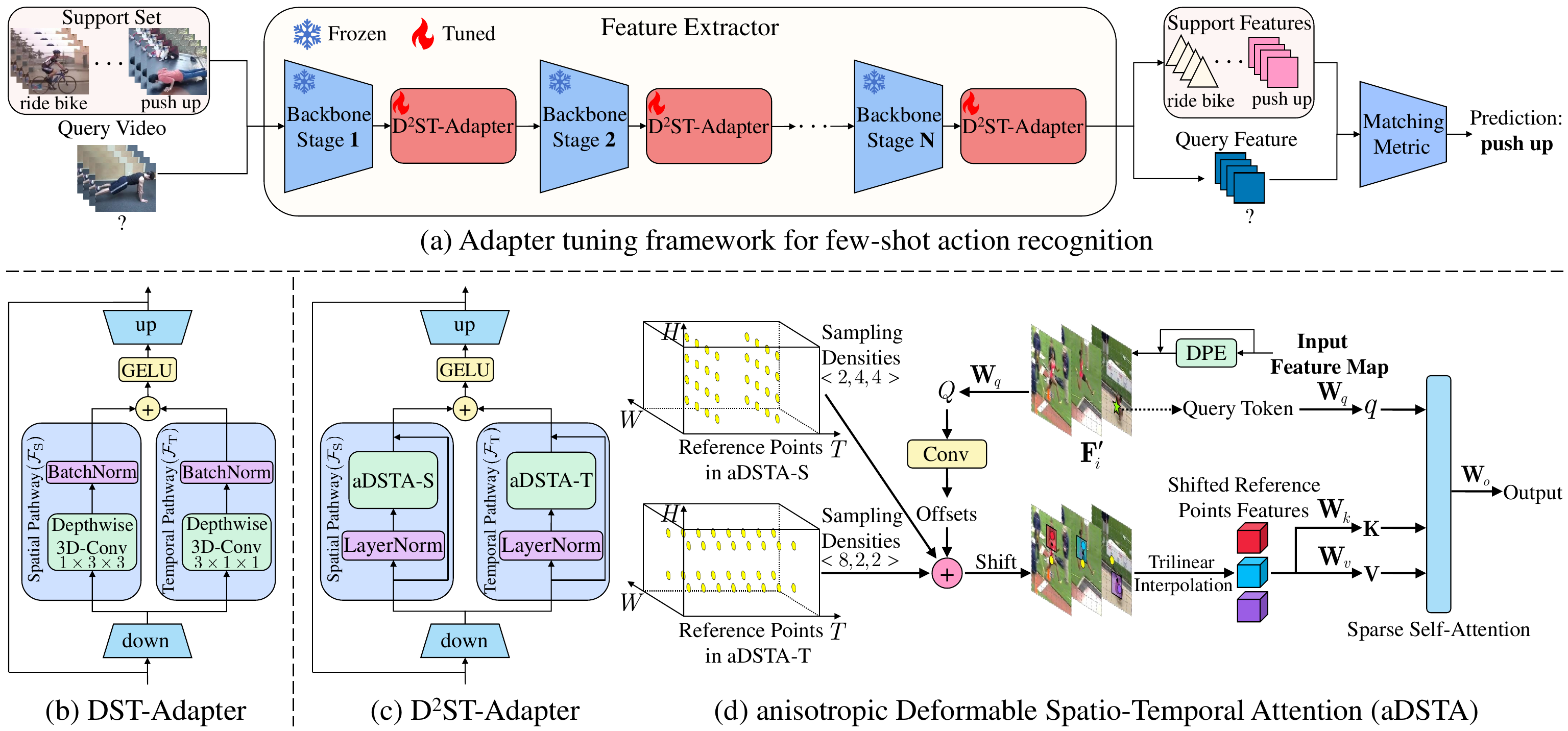}
   \vspace{-8pt}
   \caption{(a) Overall adapter tuning framework of our method. (b) DST-Adapter, a convolutional version of our model, which follows dual-pathway architecture, whereas both pathways are constructed based on 3D convolution. (c) D$^2$ST-Adapter, which is designed in a dual-pathway architecture and both pathways are modeled based on the proposed aDSTA illustrated in (d).}
   \label{fig:framework}
   \vspace{-8pt}
\end{figure*}

\noindent\textbf{Adapter Tuning Framework.}
We design a plug-and-play and lightweight adapter for video data, dubbed \emph{D$^2$ST-Adapter}, which can be integrated into most existing large image models. Thus we can leverage the powerful feature encoding capability of pre-trained models by efficient image-to-video model adaptation with only a small amount of parameter-tuning overhead. It is particularly well-suited to few-shot scenarios. Figure~\ref{fig:framework} (a) illustrates the overall adapter tuning framework of our method. Given a pre-trained feature extractor from a large model, our designed lightweight \emph{D$^2$ST-Adapter} can be selectively plugged into the middle layers of the feature extractor to perform feature adaptation. 
Only the inserted adapters are tuned during training while the pre-trained backbone keeps frozen. 
The learned features by such adapter tuning framework are further used for few-shot action recognition based on the metric-based strategies~\cite{OTAM,TRX,HyRSM}. We instantiate the backbone of feature extractor with ResNet and ViT respectively. 


\vspace{-4pt}
\subsection{D\texorpdfstring{$^2$}{}ST-Adapter}
\label{sec:adapter}
\noindent\textbf{Dual-pathway Adapter Architecture.
}
Successful action recognition from video data entails effective feature learning of both spatial semantic and temporal dynamic features. Most large vision models are typically pre-trained on image data, thus our \emph{D$^2$ST-Adapter} should be capable of capturing both the spatial features and temporal features. To this end, we design the \emph{D$^2$ST-Adapter} as a dual-pathway architecture shown in Figure~\ref{fig:framework} (c), in which the spatial pathway is responsible for capturing the spatial semantics while the temporal pathway focuses on learning the temporal dynamics. As a result, our model is able to encode the spatio-temporal features for video data in a disentangled manner.

As a common practice of typical Adapters~\cite{Adapter,ST-Adapter}, our \emph{D$^2$ST-Adapter} adopts the bottleneck architecture for reducing the computational complexity. It first downsamples the feature map into a low-dimensional feature space, then the downsampled features are fed into the spatial and temporal pathways concurrently to perform disentangled feature adaptation. Finally, both the adapted spatial and temporal features are fused by simple element-wise addition and upsampled back to the initial size. Formally, given the feature maps $\mathbf{F}_i^\text{in} \in \mathbb{R}^{T\times H\times W\times C}$ obtained from the $i$-th stage of the pre-trained backbone, containing $C$ channels of feature maps with spatial size $H\times W$ for each of $T$ frames, the feature adaptation by \emph{D$^2$ST-Adapter} can be formulated as:
\vspace{-3pt}
\begin{equation}\label{eqn:adapter}
\vspace{-3pt}
    \mathbf{F}_i^\text{out} =  \text{GELU}\left(\mathcal{F}_\text{S} (\mathbf{F}_i^\text{in} \cdot \mathbf{W}_\text{down}) \oplus \mathcal{F}_\text{T} (\mathbf{F}_i^\text{in} \cdot \mathbf{W}_\text{down})\right) \cdot \mathbf{W}_\text{up}.    
\end{equation}
Herein, $\mathbf{W}_\text{down} \in \mathbb{R}^{C \times C'}$ and $\mathbf{W}_\text{up}  \in \mathbb{R}^{C' \times C}$ are the transformation matrices of two linear layers for downsampling and upsampling, respectively. $\mathcal{F}_\text{S}$ and $\mathcal{F}_\text{T}$ denote the disentangled feature adaptation by the spatial and temporal pathways respectively. Below we discuss two feasible modeling mechanisms for $\mathcal{F}_\text{S}$ and $\mathcal{F}_\text{T}$, which leverages 3D convolution and our proposed aDSTA module, respectively.

\noindent\textbf{Modeling with 3D Convolution.}
A straightforward way to model the disentangled spatio-temporal feature adaptation ($\mathcal{F}_\text{S}$ and $\mathcal{F}_\text{T}$ in Equation~\ref{eqn:adapter}) is to employ 3D convolutional network, which can be tailored by configuring the shape of convolutional kernels to focus on learning either the spatial or the temporal features. Specifically, we model both $\mathcal{F}_\text{S}$ and $\mathcal{F}_\text{T}$ using 3D depth-wise convolutional layers followed by a batch normalization layer and a GELU layer, as shown in Figure~\ref{fig:framework} (b). The only difference between the modeling of them is that $\mathcal{F}_\text{S}$ uses $1\times 3 \times 3$ convolutional kernel for capturing spatial features while $\mathcal{F}_\text{T}$ is constructed with $3\times 1 \times 1$ convolutional layer to learn the temporal dynamics. The resulting version is called \emph{DST-Adapter}.

An important limitation of using 3D convolutions to construct \emph{D$^2$ST-Adapter}, which is also suffered by ST-Adapter, is that the convolutional operation has limited local receptive field either in the spatial or the temporal domain, especially with shallow layers for lightweight design.


\noindent\textbf{Modeling with aDSTA.}
To address the limitation of the modeling mechanism based on 3D convolution (DST-Adapter), we devise anisotropic Deformable Spatio-Temporal Attention (aDSTA) which performs feature adaptation in a global view by sparse self-attention while maintaining high computational efficiency. It adapts the deformable attention~\cite{DAT} from 2D image space to 3D spatio-temporal space. As shown in Figure~\ref{fig:framework} (d), aDSTA first samples a group of reference points with anisotropic sampling densities in the spatio-temporal space and then learns the offset for each reference point to shift them to the more informative regions. As a result, aDSTA can learn a set of informative tokens which are further used as key and value pairs shared for all queries for sparse self-attention.

We conduct several adaptations from the deformable attention in 2D space to our aDSTA in spatio-temporal 3D space. First, we employ Dynamic Position Embedding (DPE)~\cite{uniformer} implemented as a 3D depth-wise convolution layer to learn the semantic-conditioned spatio-temporal position information for each token and incorporate it into the semantic features by simple element-wise addition. Thus, the feature map $\mathbf{F}'_i \in \mathbb{R}^{T \times H  \times W \times C'}$ fused with the position information at the $i$-th stage is obtained by:
\vspace{-5pt}
\begin{equation}
\vspace{-3pt}
    \mathbf{F}'_i = \mathcal{F}_\text{DPE}(\mathbf{F}_i^\text{in} \cdot \mathbf{W}_\text{down}) \oplus (\mathbf{F}_i^\text{in} \cdot \mathbf{W}_\text{down}).
\end{equation}
Our aDSTA learns the offsets in 3D space for the reference points by a 3D convolutional network consisting of a 3D depth-wise convolution layer and a $1\times 1\times 1$ 3D convolution layer as well as a GELU in between. After shifting the reference points according to the learned offsets, we derive the features for each shifted reference point by performing trilinear interpolation among neighboring tokens within a 3D volume space around the point. For instance, the features for the shifted point $\mathbf{p}$ located at $(p_t, p_h, p_w)$ is calculated via the trilinear interpolation $\mathcal{F}_\text{tri-int}$ by:
\vspace{-5pt}
\begin{equation}\label{eq:trilinear}
\vspace{-5pt}
    \mathcal{F}_\text{tri-int}(\mathbf{p}) = \sum_{\mathbf{r}}g(p_t, r_t)\cdot g(p_h, r_h)\cdot g(p_w, r_w)\cdot \mathbf{F}'_i(\mathbf{r}),
\end{equation}
where $g(a, b)=\max(0, 1-|a-b|)$ defines an interpolating cubic volume space centered at $a$ and $\mathbf{r}=(r_t, r_h, r_w)$ indexes all tokens in the whole spatio-temporal 3D space. All shifted reference points are used as the keys and values shared for sparse self-attention. For instance, a token in $\mathbf{F}'_i$ located at $(u_t, u_h, u_w)$ serves as a query and the output $\mathbf{Z}_{i,(u_t, u_h, u_w)}$ of aDSTA is calculated by:
\vspace{-5pt}
\begin{equation}
\vspace{-3pt}
\begin{split}
    & \mathbf{q} = \mathbf{F}'_{i,(u_t, u_h, u_w)} \cdot \mathbf{W}_q, \mathbf{K} = \mathbf{P} \cdot \mathbf{W}_k, \mathbf{V} = \mathbf{P} \cdot \mathbf{W}_v, \\
    & \mathbf{Z}_{i,(u_t, u_h, u_w)} = \text{softmax}(\mathbf{q} \mathbf{K}/\sqrt{C'})\mathbf{V} \cdot \mathbf{W}_o,
\end{split}
\end{equation}
where $\mathbf{P}\in \mathbb{R}^{M\times C'}$ is feature matrix of $M$ shifted points and $\mathbf{W}_q, \mathbf{W}_k, \mathbf{W}_v, \mathbf{W}_o$ are projection matrices for the query, key, value and output respectively.

\noindent\textbf{Anisotropic sampling densities.} The learned offsets for reference points are restricted in a limited range for two reasons. First, it can prevent the potential training collapse that all reference points are shifted to the same point. Second, such restriction can ensure that each reference point can be shifted to a unique location within a local region. As a result, denser sampling of reference points typically leads to finer-grained representations. The classical deformable attention samples reference points uniformly in 2D feature space with the same sampling densities along two spatial dimensions. In contrast, we configure anisotropic sampling densities in the spatial and temporal domains. Consequently, our \emph{D$^2$ST-Adapter} can leverage aDSTA to model both spatial feature adaptation $\mathcal{F}_\text{S}$ in the spatial pathway and temporal feature adaptation $\mathcal{F}_\text{T}$ in the temporal pathway. 

We denote the sampling densities of reference points of aDSTA along the dimensions of time, height and width as $<n_t, n_s, n_s>$, thereby the total number of sampled reference points is $n_t \times n_s \times n_s$. Intuitively, an aDSTA module which samples large $n_t$ and small $n_s$ is more capable of learning the temporal features than learning the spatial features. On the other hand, sampling more reference points in the spatial domain than the temporal domain ($n_s > n_t$) generally makes aDSTA focus on learning the spatial features. Thus, we can tailor aDSTA by configuring the sampling densities to model $\mathcal{F}_\text{S}$ and $\mathcal{F}_\text{T}$ correspondingly: 
\begin{itemize}[leftmargin =*, itemsep = 2pt, topsep = 0pt]
    \item \textbf{aDSTA-S} for modeling $\mathcal{F}_\text{S}$ in the spatial pathway, which is sampled  with larger $n_s$ and smaller $n_t$ ($n_s > n_t$).
    \item \textbf{aDSTA-T} for modeling $\mathcal{F}_\text{T}$ in the temporal pathway, which samples denser reference points in temporal domain than in spatial domain ($n_t > n_s$).
\end{itemize}
The values of $n_t$ and $n_s$ are tuned as hyper-parameters. 

\vspace{-4pt}
\subsection{End-to-End Adapter Tuning}
The learned features by our adapter tuning framework are further used for few-shot action recognition based on the metric-based strategy. Following ProtoNet~\cite{protonet}, it first calculates the frame-wise $L_2$ distance matrix between the query video and each class prototype derived by averaging the corresponding support samples. Then the distance matrix is used to calculate the matching similarity between the query and each class for prediction. Formally, the similarity between query $q$ and class prototype $c$ can be expressed as:
\vspace{-3pt}
\begin{equation}
\vspace{-3pt}
    s(\mathbf{F}_q, \mathbf{F}_c)=\mathbf{M}([\mathbf{F}_q^1, \cdots, \mathbf{F}_q^T], [\mathbf{F}_c^1, \cdots, \mathbf{F}_c^T]),
\end{equation}
where $\mathbf{F}_q^i$ and $\mathbf{F}_c^j$ denote the features of the $i$-th frame in query $q$ and the $j$-th frame in class prototype $c$, and $\mathbf{M}$ is the matching metric such as Bi-MHM~\cite{HyRSM}. During training, we take the similarities for each class as logits and optimize the model in an end-to-end manner using the same cross-entropy loss as in~\cite{OTAM,TRX,HyRSM}. For inference, we classify the query as the support class closest to it.
\section{Experiments}
\subsection{Experimental Setup}

\noindent\textbf{Datasets.} We conduct experiments on five standard few-shot action recognition benchmarks, including SSv2-Full~\cite{SSV2}, SSv2-Small~\cite{SSV2}, Kinetics~\cite{Kinetics}, HMDB51~\cite{HMDB51}, and UCF101~\cite{UCF101}.
Following the typical data split~\cite{CMN, OTAM, MoLo}, for SSv2-Full, SSv2-Small and Kinetics, we select 64/12/24 classes from the datasets as the training/validation/test set, respectively. As for HMDB51 and UCF101, we adopt the same data split as~\cite{ARN, MoLo}.

\begin{table*}[t]
\centering
\caption{Classification accuracy ($\%$) comparison on 5 datasets, under 5-way 1-shot and 5-way 5-shot settings. The highest results are highlighted in \textbf{bold} and the second best results are \underline{underlined}. ``PEFT'' indicates parameter-efficient fine-tuning. ``$*$'' denotes multi-modal methods incorporating additional text input and text encoder.
}
\vspace{-5pt}
\resizebox{\linewidth}{!}{
\begin{tabular}{l|c|c|cc|cc|cc|cc|cc}
\toprule
\multicolumn{1}{l|}{\multirow{2}{*}{Method}} & \multicolumn{1}{c|}{\multirow{2}{*}{Pre-training}} &
\multicolumn{1}{c|}{\multirow{2}{*}{Fine-tuning}} & \multicolumn{2}{c|}{{SSv2-Full}} & \multicolumn{2}{c|}{{SSv2-Small}} &
\multicolumn{2}{c|}{{Kinetics}} & \multicolumn{2}{c|}{{HMDB51}} & \multicolumn{2}{c}{{UCF101}} \\
\multicolumn{1}{l|}{} & \multicolumn{1}{c|}{} & \multicolumn{1}{c|}{} &
\multicolumn{1}{l}{1-shot} & 5-shot & \multicolumn{1}{l}{1-shot} & 5-shot & \multicolumn{1}{l}{1-shot} & 5-shot &
\multicolumn{1}{l}{1-shot} & 5-shot & \multicolumn{1}{l}{1-shot} & 5-shot \\
\midrule
CMN~\cite{CMN}          & INet-RN50 & Full
& 34.4  & 43.8  & 33.4  & 46.5  & 57.3  & 76.0  & $-$   & $-$   & $-$   & $-$  \\ 
OTAM~\cite{OTAM}        & INet-RN50 & Full
& 42.8  & 52.3  & 36.4  & 48.0  & 73.0  & 85.8  & 54.5  & 68.0  & 79.9  & 88.9 \\
AMeFu-Net~\cite{AMeFu}  & INet-RN50 & Full
& $-$   & $-$   & $-$   & $-$   & 74.1  & 86.8  & 60.2  & 75.5  & 85.1  & 95.5 \\
ITANet~\cite{ITANet}    & INet-RN50 & Full
& 49.2  & 62.3  & 39.8  & 53.7  & 73.6  & 84.3  & $-$   & $-$   & $-$   & $-$  \\
TRX~\cite{TRX}          & INet-RN50 & Full
& 42.0  & 64.6  & 36.0  & 59.1  & 63.6  & 85.9  & 53.1  & 75.6  & 78.2  & 96.1 \\
TA$^{2}$N~\cite{TA2N}   & INet-RN50 & Full
& 47.6  & 61.0  & $-$   & $-$   & 72.8  & 85.8  & 59.7  & 73.9  & 81.9  & 95.1 \\
MTFAN~\cite{MTFAN}      & INet-RN50 & Full
& 45.7  & 60.4  & $-$   & $-$   & 74.6  & 87.4 & 59.0  & 74.6  & 84.8  & 95.1 \\
STRM~\cite{STRM}        & INet-RN50 & Full
& 43.1  & 68.1  & 37.1  & 55.3  & 62.9  & 86.7  & 52.3  & 77.3  & 80.5  & \underline{96.9} \\
HyRSM~\cite{HyRSM}      & INet-RN50 & Full
& 54.3  & 69.0  & 40.6  & 56.1  & 73.7  & 86.1  & 60.3  & 76.0  & 83.9  & 94.7 \\
Nguyen~\emph{et al.}~\cite{nguyen}  & INet-RN50 & Full
& 43.8  & 61.1  & $-$   & $-$   & 74.3  & 87.4 & 59.6  & 76.9  & 84.9  & 95.9 \\
Huang~\emph{et al.}~\cite{huang}    & INet-RN50 & Full
& 49.3  & 66.7  & 38.9  & \textbf{61.6} & 73.3  & 86.4  & 60.1  & 77.0  & 71.4  & 91.0 \\ 
HCL~\cite{HCL}  & INet-RN50 & Full
& 47.3  & 64.9  & 38.7  & 55.4  & 73.7  & 85.8  & 59.1  & 76.3  & 82.5  & 93.9 \\
SA-CT~\cite{SA-CT}      & INet-RN50 & Full
& 48.9  & 69.1  & $-$   & $-$   & 71.9  & 87.1  & 60.4  & \textbf{78.3} & 85.4  & 96.4 \\ 
SloshNet~\cite{sloshnet}& INet-RN50 & Full
& 46.5  & 68.3  & $-$   & $-$   & 70.4  & 87.0  & 59.4  & \underline{77.5} & 86.0 & \textbf{97.1}  \\
GgHM~\cite{GgHM}        & INet-RN50 & Full
& 54.5  & 69.2  & $-$   & $-$   & 74.9 & 87.4 & 61.2 & 76.9 & 85.2 & 96.3 \\
MoLo~\cite{MoLo}        & INet-RN50 & Full
& \underline{56.6} & 70.6 & \underline{42.7}  & 56.4  & 74.0  & 85.6  & 60.8  & 77.4  & 86.0  & 95.5 \\
Zheng~\emph{et al.}~\cite{zheng2024saliency} & INet-RN50 & Full
& 55.4  & \underline{71.4}  & 41.3  & 57.5  & \underline{75.6}  & \underline{87.6}  & \underline{61.5}  & 76.0  & \textbf{87.8}  & 96.0 \\
ST-Adapter~\cite{ST-Adapter}    & INet-RN50 & PEFT
& 52.2  & 68.7  & 41.9  & 55.7  & 73.0  & 85.1  & 60.3  & 74.7  & 84.6  & 94.5 \\
\textbf{D$^2$ST-Adapter} & INet-RN50 & PEFT
& \textbf{57.0} & \textbf{73.6} & \textbf{45.8} & \underline{60.9} & \textbf{75.8} & \textbf{87.7} & \textbf{61.6} & 76.6 & \underline{86.9} & 95.6 \\
\midrule
AIM~\cite{AIM}  & CLIP-ViT-B & PEFT
& 63.7  & 79.2  & 52.8  & 67.5  & 88.4  & 95.3  & 74.2  & 86.9  & 95.4  & 98.5 \\
DUALPATH~\cite{DUALPATH}  & CLIP-ViT-B & PEFT
& \underline{64.5}  & \underline{79.8}  & \underline{53.5}  & \underline{68.1}  & \underline{88.8}
& \underline{95.4}  & \underline{74.9}  & \underline{87.5}  & 95.7  & 98.7 \\
ST-Adapter~\cite{ST-Adapter} & CLIP-ViT-B & PEFT
& 64.2  & 79.5  & 53.1  & 68.0  & 88.5  & 95.1  & 74.1  & 87.3  & \underline{95.9}  & \underline{98.9} \\
\textbf{D$^2$ST-Adapter} & CLIP-ViT-B & PEFT
& \textbf{66.7} & \textbf{81.9} & \textbf{55.0} & \textbf{69.3} & \textbf{89.3}
& \textbf{95.5} & \textbf{77.1} & \textbf{88.2} & \textbf{96.4} & \textbf{99.1} \\
\midrule
CLIP-FSAR~\cite{CLIP-FSAR}$^{*}$ & CLIP-ViT-B & Full
& 62.1 & 72.1 & 54.6 & 61.8 & 94.8 & 95.4 & 77.1 & 87.7 & 97.0 & 99.1 \\
EMP-Net~\cite{EMP}$^{*}$ & CLIP-ViT-B & PEFT
& 63.1 & 73.0 & 57.1 & 65.7 & 89.1 & 93.5 & 76.8 & 85.8 & 94.3 & 98.2 \\
MA-FSAR~\cite{MA-FSAR}$^{*}$ & CLIP-ViT-B & PEFT
& 63.3 & 72.3 & 59.1 & 64.5 & \underline{95.7} & 96.0 & 83.4 & 87.9 & 97.2 & 99.2 \\
Task-Adapter~\cite{Task-Adapter}$^{*}$ & CLIP-ViT-B & PEFT
& \underline{71.3} & \underline{74.2} & \underline{60.2} & \underline{70.2} & 95.0
& \underline{96.8} & \underline{83.6} & \underline{88.8} & \underline{98.0} & \underline{99.4} \\
\textbf{D$^2$ST-Adapter-MM}$^{*}$ & CLIP-ViT-B & PEFT
& \textbf{74.1} & \textbf{83.6} & \textbf{63.6} & \textbf{71.9} & \textbf{96.2}
& \textbf{97.2} & \textbf{84.3} & \textbf{89.0} & \textbf{98.3} & \textbf{99.5} \\
\bottomrule
\end{tabular}
}
\vspace{-10pt}
\label{table_sota}
\end{table*}

\noindent\textbf{Implementation details.}
To evaluate \emph{D$^2$ST-Adapter}, we instantiate our method with both ResNet-50~\cite{Resnet} and CLIP-ViT-B/16~\cite{ViT}. We insert one \emph{D$^2$ST-Adapter} module in each stage of the pre-trained model (4 stages for ResNet-50 and 12 stages for CLIP-ViT-B/16), as shown in Figure~\ref{fig:framework} (a). For a fair comparison with previous methods~\cite{HyRSM,MoLo}, we uniformly sample 8 frames (\ie $T=8$) as the input of a video. Bi-MHM~\cite{HyRSM} is used as the matching metric in comparison with state-of-the-art methods. Besides, OTAM~\cite{OTAM} and TRX~\cite{TRX} are additionally employed in the ablation study. We report the average classification accuracy of 10,000 episodes randomly selected from the test set. Other implementation details follow HyRSM~\cite{HyRSM}, which are provided in the supplementary material.

\subsection{Comparison with State-of-the-Art Methods}
We compare our method with other methods on two types of datasets: temporal-related benchmarks including SSv2-Full and SSv2-Small datasets, and spatial-related benchmarks inlcuding Kinetics, HMDB51, and UCF101 datasets.

\noindent\textbf{Benchmarks sensitive to temporal features.}
Table~\ref{table_sota} shows the comparative results of different methods on two temporal-related datasets, \ie SSv2-Full and SSv2-Small, which are quite challenging since both the temporal and spatial features are crucial to the action recognition. Our \emph{D$^2$ST-Adapter} achieves the best performance and outperforms other methods by a large margin in most few-shot settings. These results demonstrate the advantages of our method over other methods in dealing with challenging scenarios where the temporal features are critical for action recognition. In particular, our \emph{D$^2$ST-Adapter} substantially outperforms all adapter tuning methods for video data including AIM, DUALPATH, and ST-Adapter, highlighting the advantages of its core design, namely built-in disentangled adaptation of spatial and temporal features.

\noindent\textbf{Benchmarks relying on spatial features.}
Experimental results on the spatial-related benchmarks including Kinetics, HMDB51, and UCF101, are also presented in Table~\ref{table_sota}. Our method still performs best in most settings, which manifest the effectiveness and robustness of our method in these scenarios. Besides, the performance improvement by our \emph{D$^2$ST-Adapter} over other method, especially ST-Adapter, is smaller than that on SSv2 datasets. It is reasonable since the action recognition in these datasets relies less on the temporal features and thus the disentangled feature encoding of our method yields limited performance gain.

\begin{table}[t]
\centering
\caption{Comparison with previous adapter-based methods on traditional full-shot action recognition task with CLIP-ViT-B/16 backbone. The testing views (\# frames per clip $\times$ \# temporal clip $\times$ \# spatial crop) are set as 8$\times$1$\times$3 for K400 and 8$\times$3$\times$1 for SSv2.}
\resizebox{0.6\linewidth}{!}{
\begin{tabular}{l|c|c}
\toprule
Method & K400 & SSv2 \\
\midrule
AIM~\cite{AIM}                  & 83.9 & 66.4 \\
DUALPATH~\cite{DUALPATH}        & 84.1 & 67.3 \\
ST-Adapter~\cite{ST-Adapter}    & 82.0 & 67.1 \\
\textbf{D$^2$ST-Adapter} (Ours) & \textbf{84.5} & \textbf{68.0} \\
\bottomrule
\end{tabular}
}
\vspace{-10pt}
\label{table_traditional}
\end{table}

\noindent\textbf{Comparison with multimodal methods.} To have a fair comparison with the multimodal methods for few-shot action recognition, we extend our model to multi-modality (\emph{D$^2$ST-Adapter-MM}). This extension draws on the multimodal framework of CLIP-FSAR~\cite{CLIP-FSAR} to align with other methods. Table~\ref{table_sota} shows that our method outperforms all other multimodal methods, especially on SSv2-Full and SSv2-Small datasets. These findings consistently highlight the effectiveness and robustness of our \emph{D$^2$ST-Adapter}.

\begin{figure}[t]
  \centering
   \includegraphics[width=\linewidth]{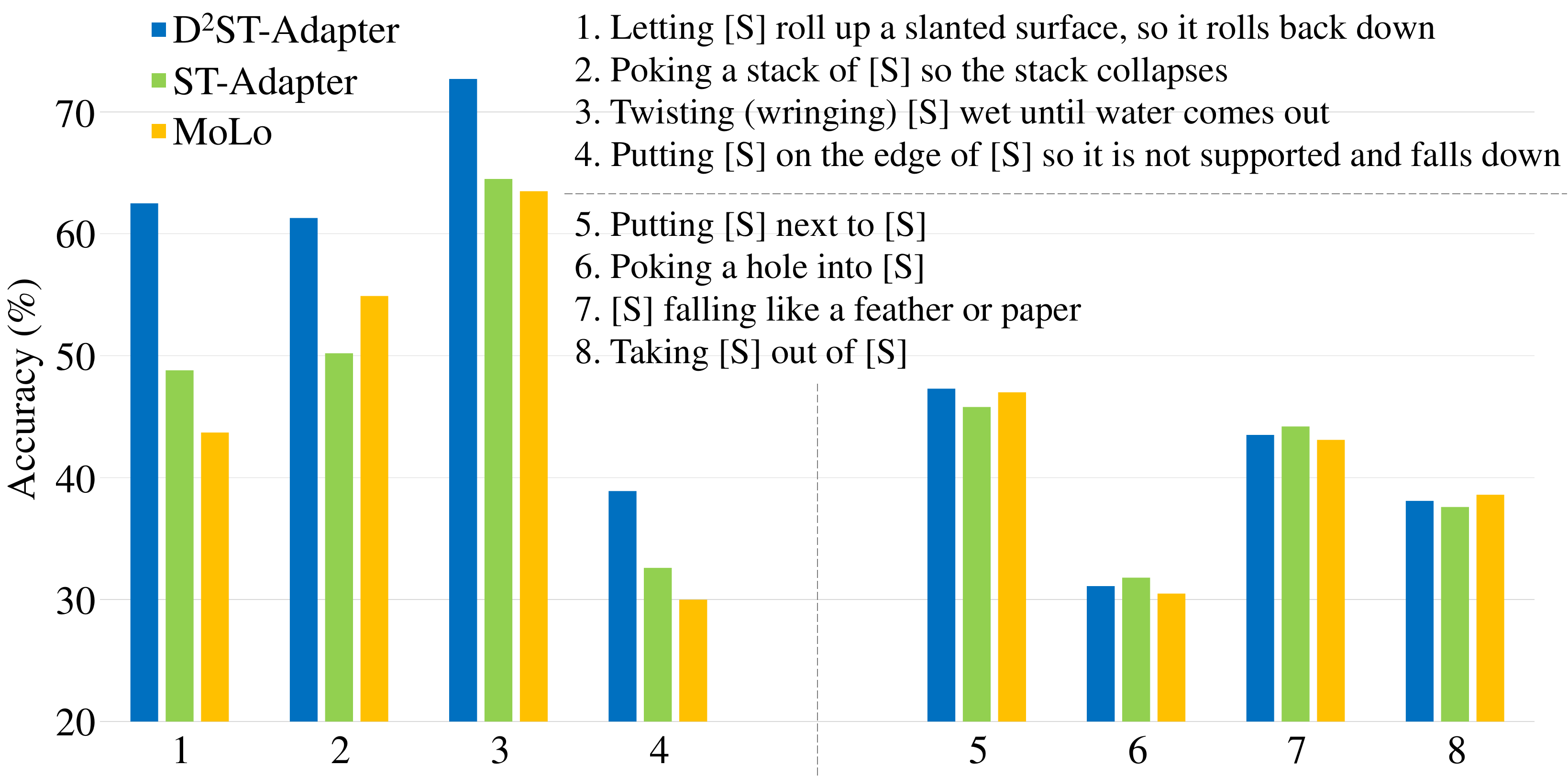}
   \vspace{-18pt}
   \caption{Illustration of actions our method excels at (1-4) and performs poorly at (5-8), respectively. The performance is evaluated on SSv2-Small benchmark in 1-shot setting.}
   \vspace{-8pt}
   \label{fig:Class}
\end{figure}

\noindent\textbf{Comparison on traditional (full-shot) action recognition task.}
Our model is particularly effective in the few-shot scenario while it can also be applied to traditional action recognition tasks using full training samples. To obtain a more comprehensive assessment of our model, we conduct experiments on two traditional action recognition benchmarks, Kinetics-400 (K400)~\cite{Kinetics} and Something-Something-v2 (SSv2)~\cite{SSV2}, using the same settings as ST-Adapter~\cite{ST-Adapter}. Table~\ref{table_traditional} shows that our method still outperforms these adapter-based methods on both benchmarks.

\begin{figure}[t]
  \centering
   \includegraphics[width=0.7\linewidth]{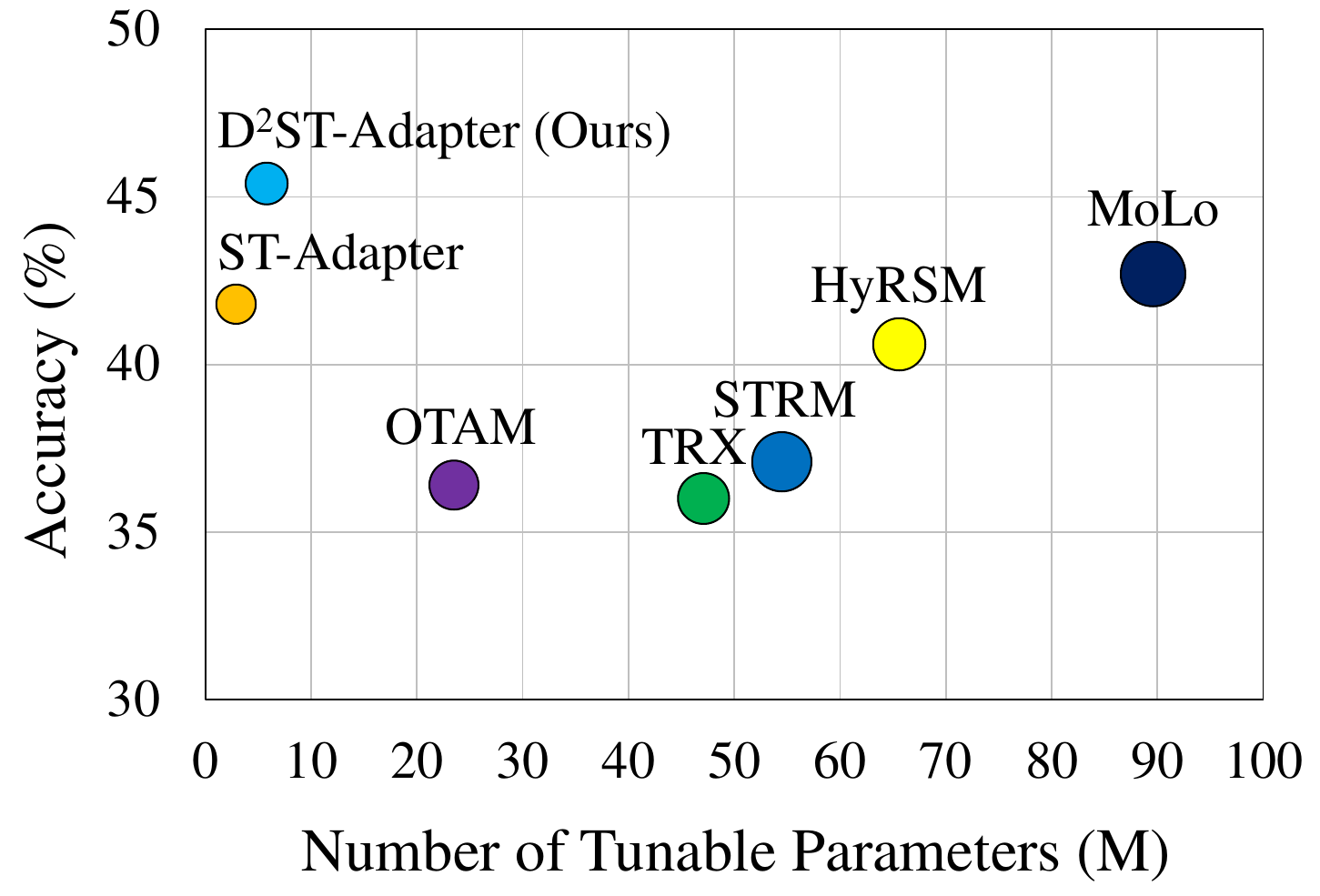}
   \vspace{-6pt}
   \caption{Performance and efficiency of different methods with ResNet-50 backbone on SSv2-Small benchmark in 1-shot setting. Circle size indicates the memory usage.}
   \vspace{-6pt}
   \label{fig:efficiency}
\end{figure}

\noindent\textbf{Case Study of the Pros and Cons of \emph{D\texorpdfstring{$^2$}{}ST-Adapter}.}
To gain more insights into the strengths and weaknesses of our \emph{D$^2$ST-Adapter}, we select four categories of actions where our method excels at, as well as four categories where our model struggles in comparison to the state-of-the-art MoLo and ST-Adapter. Figure~\ref{fig:Class} shows that our model achieves large performance superiority over other methods on the relatively complex actions requiring careful reasoning via learning temporal features for recognition, whilst performing on par with other methods on simple actions that can be recognized primarily based on spatial features. Such results are consistent with the conclusion that our model excels at recognizing actions involving complex temporal dynamics.

\begin{table}[t]
\centering
\caption{Comparison with other adapter-based methods using CLIP-ViT-B/16 as backbone on efficiency. Accuracy on SSv2-Full benchmark in 1-shot setting is also provided.}
\vspace{-5pt}
\resizebox{\linewidth}{!}{
\begin{tabular}{l|c|c|c|c|c}
\toprule
\multirow{2}{*}{Method} & Tunable & Memory & \multirow{2}{*}{GFLOPs} & \multirow{2}{*}{Latency} & \multirow{2}{*}{Acc.} \\
 & Params & Usage & & & \\ 
\midrule
AIM~\cite{AIM}               & 14.3 M & 19.0 GB & 208 & 14.1 ms & 63.7 \\
DUALPATH~\cite{DUALPATH}     & 14.3 M & 16.7 GB & 171 & 11.6 ms & 64.5 \\
ST-Adapter~\cite{ST-Adapter} & 14.3 M & 16.1 GB & 163 & 12.6 ms & 64.2 \\
\textbf{D$^2$ST-Adapter}     & 7.3 M  & 17.3 GB & 150 & 18.2 ms & \textbf{66.7} \\
\bottomrule
\end{tabular}
}
\vspace{-12pt}
\label{table_efficiency2}
\end{table}

\noindent\textbf{Comparison of efficiency.}
We compare the efficiency between our \emph{D$^2$ST-Adapter} and other methods in terms of tunable parameter size and memory usage in Figure~\ref{fig:efficiency}. It shows that the tunable parameter size of both our model and ST-Adapter is significantly smaller than other methods that are based on full fine-tuning, which reveals the advantage of the adapter tuning paradigm over the full fine-tuning paradigm for task adaptation. Meanwhile, our method outperforms these full fine-tuning based methods substantially with much lower parameter-finetuning overhead, which validates the superiority of our method.

We further compare our model with other adapter-based methods on efficiency in Table~\ref{table_efficiency2}. Our \emph{D$^2$ST-Adapter} performs best while maintaining relatively high efficiency due to the lightweight design of the proposed aDSTA module and the smaller number of inserted adapters (only one per block in our method \vs two or four per block in other adapter-based methods~\cite{ST-Adapter,AIM,DUALPATH}).

\begin{figure*}[t]
  \centering
   \includegraphics[width=\linewidth]{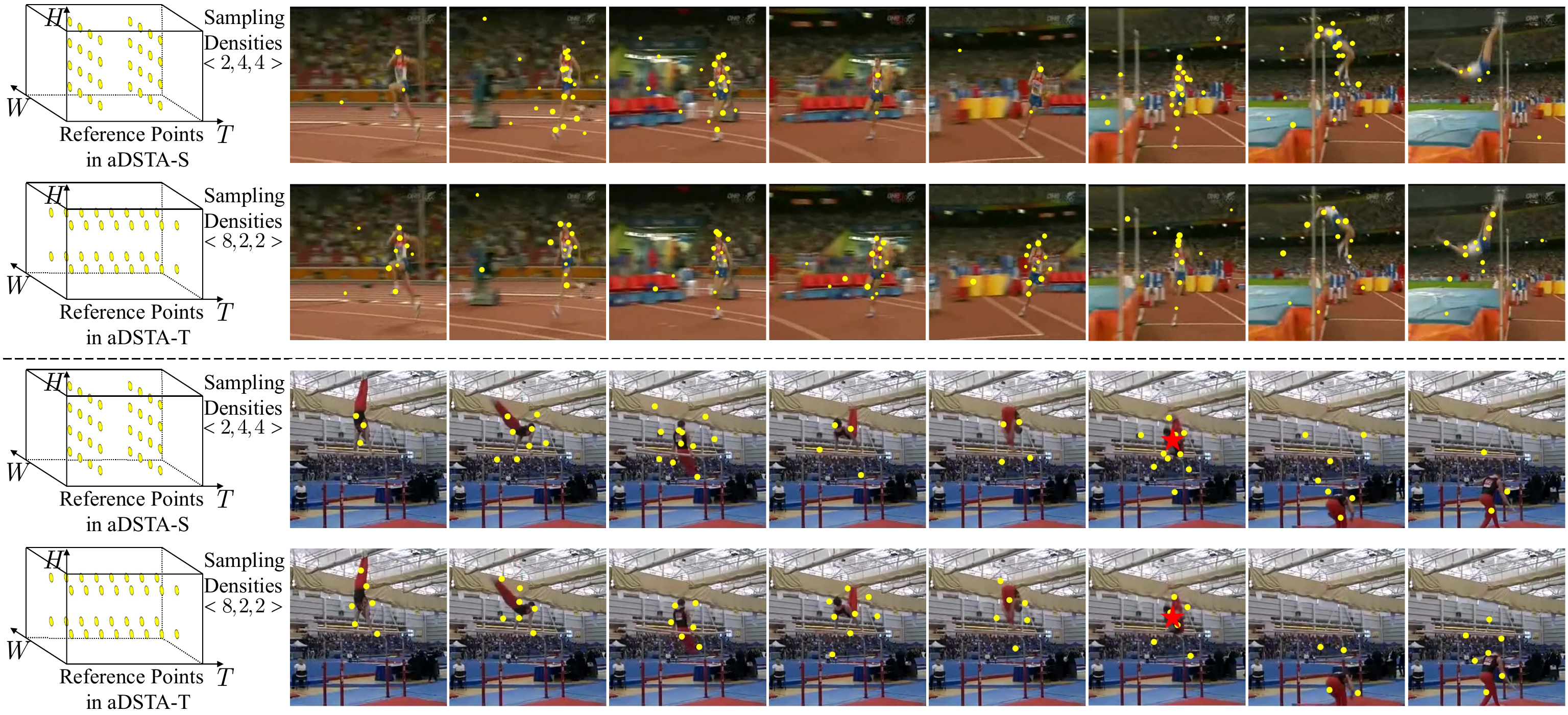}
   \vspace{-16pt}
   \caption{Visualization of important shifted reference points \emph{w.r.t} all queries (first sample), and important shifted reference points \emph{w.r.t} a given query, marked with a red star (second sample). More visualization results are provided in the supplementary material.}
   \vspace{-10pt}
   \label{fig:visualization}
\end{figure*}

\begin{table}[t]
\centering
\caption{Comparison between different adapters.}
\vspace{-7pt}
\resizebox{0.85\linewidth}{!}{
\begin{tabular}{l|cc|cc}
\toprule
\multirow{2}{*}{Method} & \multicolumn{2}{c|}{{SSv2-Full}} & \multicolumn{2}{c}{{Kinetics}} \\
 & 1-shot & 5-shot & 1-shot & 5-shot \\
\midrule
Full Fine-tuning   & 44.9 & 57.0 & 72.4 & 84.3 \\
Vanilla-Adapter    & 45.3 & 57.6 & 72.6 & 84.8 \\
ST-Adapter         & 52.2 & 68.7 & 73.0 & 85.1 \\
DST-Adapter (Ours) & 53.7 & 69.6 & 74.4 & 86.4 \\
\textbf{D$^2$ST-Adapter} (Ours)    & \textbf{57.0} & \textbf{73.6} & \textbf{75.8} & \textbf{87.7} \\
\bottomrule
\end{tabular}
}
\vspace{-4pt}
\label{table_adapter}
\end{table}

\begin{table}[t]
\centering
\caption{Effect of the anisotropic sampling of aDSTA.}
\vspace{-7pt}
\resizebox{\linewidth}{!}{
\begin{tabular}{l|l|cc|cc}
\toprule
\multirow{2}{*}{Method} & \multirow{2}{*}{Configurations} & \multicolumn{2}{c|}{{SSv2-Full}} & \multicolumn{2}{c}{{Kinetics}} \\
 & & 1-shot & 5-shot & 1-shot & 5-shot \\
\midrule
 & aDSTA-S & 55.2 & 71.9 & 74.1 & 86.0 \\
Single branch & aDSTA-T & 56.3 & 72.5 & 73.8 & 85.8 \\
 & aDSTA-Uniform & 55.8 & 72.2 & 73.9 & 86.0 \\
\midrule
Dual branches & aDSTA-S \& aDSTA-T  & \textbf{57.0} & \textbf{73.6} & \textbf{75.8} & \textbf{87.7} \\
\bottomrule
\end{tabular}
}
\vspace{-10pt}
\label{table_sampling_densities}
\end{table}

\vspace{-2pt}
\subsection{Ablation study}
We conduct ablation study on SSv2-Full and Kinetics datasets, using ResNet-50 as the pre-trained backbone and Bi-MHM~\cite{HyRSM} as the matching metric.

\noindent\textbf{Comparison between different Adapters.} We first compare the performance of different Adapters in Table~\ref{table_adapter}. Vanilla-Adapter~\cite{Adapter} only contains the downsampling and upsampling layers with a GELU nonlinearity in between. As described in Section `Method', DST-Adapter is the convolutional version of our \emph{D$^2$ST-Adapter} which models both the spatial and temporal pathways with 3D convolution. The performance of full fine-tuning is also provided for reference. We make following observations. 
\begin{itemize}[leftmargin =*, itemsep = 0pt, topsep = 0pt]
    \item Effect of learning temporal features by adapters. ST-Adapter outperforms Vanilla-Adapter distinctly, which indicates that using 3D convolution to learn joint spatio-temporal features facilitates the feature adaptation. 
    \item Effect of disentangled encoding of the spatial and temporal features is demonstrated by the performance improvement from ST-Adapter to DST-Adapter benefited from the dual-pathway adapter architecture. 
    \item Effect of the proposed aDSTA. Comparing our \emph{D$^2$ST-Adapter} with DST-Adapter, the proposed aDSTA yields a large performance gain, especially on SSv2-Full which demands effective encoding of temporal features.
\end{itemize}

\noindent\textbf{Effect of anisotropic sampling of aDSTA.} To investigate the effect of proposed anisotropic sampling scheme of aDSTA, we compare the performance of our \emph{D$^2$ST-Adapter} using four different configurations of the aDSTA: 1) only `aDSTA-S' modeling the spatial pathway is used; 2) only `aDSTA-T' for the temporal pathway is preserved; 3) `aDSTA-Uniform'  performs sampling with uniform density in both the spatial and temporal domains; and 4) `aDSTA-S \& aDSTA-T' corresponds the fully functional \emph{D$^2$ST-Adapter}. 
Table~\ref{table_sampling_densities} shows that the distinct performance gap between these two settings reveals the advantage of configuring specialized sampling densities for the spatial and temporal pathways.

\begin{table}[t]
\centering
\caption{Evaluation of generalization of our \emph{D$^2$ST-Adapter} across different matching metrics on SSv2-Full and Kinetics. The state-of-the-art methods for each metric are involved into comparison.}
\vspace{-6pt}
\resizebox{\linewidth}{!}{
\begin{tabular}{c|c|cc|cc}
\toprule
\multirow{2}{*}{Matching Metric} & \multirow{2}{*}{Method} & \multicolumn{2}{c|}{{SSv2-Full}} & \multicolumn{2}{c}{{Kinetics}} \\
 & & 1-shot & 5-shot & 1-shot & 5-shot \\
\midrule
                    & RFPL    & 47.0 & 61.0 & 74.6 & \textbf{86.8} \\
OTAM                & MoLo    & 55.0 & 69.6 & 73.8 & 85.1 \\
                    & \textbf{D$^2$ST-Adapter} & \textbf{56.0} & \textbf{72.8} & \textbf{74.7} & \textbf{86.8} \\
\midrule
                    & RFPL    & 44.6 & 64.6 & 66.2 & 87.3 \\
TRX                 & MoLo    & 45.6 & 66.1 & 64.8 & 86.3 \\
                    & \textbf{D$^2$ST-Adapter} & \textbf{47.7} & \textbf{68.6} & \textbf{66.4} & \textbf{87.6} \\
\midrule
                    & HyRSM   & 54.3 & 69.0 & 73.7 & 86.1 \\
Bi-MHM              & MoLo    & 56.6 & 70.6 & 74.0 & 85.6 \\
                    & \textbf{D$^2$ST-Adapter} & \textbf{57.0} & \textbf{73.6} & \textbf{75.8} & \textbf{87.7} \\
\bottomrule
\end{tabular}
}
\vspace{-14pt}
\label{table_matching_metrics}
\end{table}

\noindent\textbf{Generalization Across Different Matching Metrics.}
To evaluate the generalization of our \emph{D$^2$ST-Adapter} across different matching metrics, we conduct the experiments using three classical matching metrics on SSv2-Full and Kinetics, including OTAM~\cite{OTAM}, TRX~\cite{TRX}, and Bi-MHM~\cite{HyRSM}, and compare our model with the state-of-the-art methods corresponding to each metric. Table~\ref{table_matching_metrics} presents the results, which show that our \emph{D$^2$ST-Adapter} consistently outperforms other methods, especially on the challenging SSv2-Full benchmark, validating the well generalization of the proposed \emph{D$^2$ST-Adapter} across different matching metrics.

\noindent\textbf{Visualization.}
We visualize the shifted reference points learned by our \emph{D$^2$ST-Adapter} to investigate the proposed aDSTA qualitatively. Following Deformable Attention~\cite{DAT}, for each shifted reference point serving as key and value, we accumulate the attention weights assigned by all queries as its relative importance. Then we visualize top-100 most important shifted reference points in each pathway, with circle size indicating their importance, as shown in the first sample of Figure~\ref{fig:visualization}. We also provide visualization of the most relevant reference points \emph{w.r.t} a certain salient region (token) in the second sample of Figure~\ref{fig:visualization}. To be specific, we manually select a salient region (indicated by a red star marker) as the query token and visualize top-50 relevant shifted reference points to the query in terms of attention weights.

The results show that the shifted reference points learned by our model can always focus on the salient targets that are critical for action recognition in both the spatial and temporal pathways. Another observation is that the shifted points in the spatial pathway are distributed densely in some specific frames to capture the spatial appearance features for the targets, while the shifted points in the temporal pathway are distributed uniformly among all frames to capture the temporal dynamics. This is consistent with the anisotropic sampling of aDSTA for different pathways. 
\section{Concluding Remarks}
We present \emph{D$^2$ST-Adapter}, which is a novel adapter tuning method for few-shot action recognition. It is designed in a dual-pathway architecture, which allows for disentangled encoding of spatial and temporal features. Moreover, we design the aDSTA module, which enables \emph{D$^2$ST-Adapter} to encode features in a global view while maintaining lightweight design. Extensive experiments demonstrate the superiority of our model over state-of-the-art methods, especially in challenging scenarios involving complex temporal features.

{
    \small
    \bibliographystyle{ieeenat_fullname}
    \bibliography{main}
}

\clearpage
\setcounter{page}{1}
\maketitlesupplementary
\appendix

\section{Effect of the Sampling Densities of \emph{D\texorpdfstring{$^2$}{}ST-Adapter}}
We compare the efficiency and performance of different sampling densities in 1-shot setting in Table~\ref{table_sampling_densities_ablation}. It demonstrates that sampling denser reference points yields marginal performance gain at the expense of increased memory usage. Our model is set to $<$2, 4, 4$>$ $\&$ $<$8, 2, 2$>$ to strike a good balance between efficiency and performance.

\section{Effect of the Number of Input Frames}
To analyze the impact of the number of input frames, we compare the SSv2-Full 1-shot performance when using different number of frames as input under two sampling densities settings. As shown in Table~\ref{table_input_frames_ablation}, more input frames and denser sampling both yield better performance while incurring more computational cost.

\section{Instantiation on More Backbones}
To investigate the performance of our \emph{D$^2$ST-Adapter} on smaller backbones, we instantiate our model with ResNet-18 and ResNet-34 pre-trained on ImageNet, and conduct experiments to compare our model with other methods under the same experimental settings. Table~\ref{table_smaller_backbone} shows that our \emph{D$^2$ST-Adapter} achieves the best performance using both backbones in all settings, which demonstrates the robustness of our method across different backbones. While DST-Adapter, the convolutional version of our model, also performs well, it is still inferior to \emph{D$^2$ST-Adapter}, which manifests the effectiveness of the proposed anisotropic Deformable Spatio-Temporal Attention (aDSTA).

For larger backbones, we conduct experiments on CLIP-ViT-L/14, which has a total of 303.2 million parameters, significantly larger than CLIP-ViT-B/16, which has a total of 85.8 million parameters. The results in Table~\ref{table_larger_backbone} show that our model can still consistently outperform ST-Adapter and Full Fine-tuning on the larger and stronger backbone.

\begin{table}[t]
\centering
\caption{Effect of the sampling densities of \emph{D$^2$ST-Adapter} in CLIP-ViT-B.}
\resizebox{\linewidth}{!}{
\begin{tabular}{c|c|c|c}
\toprule
Sampling Densities & \multirow{2}{*}{Memory Usage} & \multirow{2}{*}{SSv2-Full} & \multirow{2}{*}{Kinetics} \\
(Spatial \& Temporal) & & \\
\midrule
$<$1, 4, 4$>$ $\&$ $<$4, 2, 2$>$  & 17.0 GB & 65.9 & 89.0 \\
$<$2, 4, 4$>$ $\&$ $<$8, 2, 2$>$  & 17.3 GB & 66.7 & 89.3 \\
$<$2, 6, 6$>$ $\&$ $<$8, 3, 3$>$  & 18.0 GB & 66.9 & 89.2 \\
$<$2, 8, 8$>$ $\&$ $<$8, 4, 4$>$  & 18.8 GB & 66.9 & 89.6 \\
$<$4, 8, 8$>$ $\&$ $<$16, 4, 4$>$ & 20.3 GB & 67.2 & 89.3 \\
\bottomrule
\end{tabular}
}
\label{table_sampling_densities_ablation}
\end{table}

\begin{table}[t]
\centering
\caption{Effect of the number of input frames in SSv2-Full 1-shot setting with CLIP-ViT-B backbone.}
\resizebox{0.7\linewidth}{!}{
\begin{tabular}{c|ccc}
\toprule
Sampling Densities & \multicolumn{3}{c}{{Sampling Frames}} \\
(Spatial \& Temporal) & 4 & 8 & 16 \\
\midrule
$<$1, 4, 4$>$ $\&$ $<$4, 2, 2$>$ & 63.1 & 65.9 & 69.0 \\
$<$2, 4, 4$>$ $\&$ $<$8, 2, 2$>$ & 63.6 & 66.7 & 69.9 \\
\bottomrule
\end{tabular}
}
\label{table_input_frames_ablation}
\end{table}

\begin{table}[t]
\centering
\caption{Effect of the inserted position of \emph{D$^2$ST-Adapter} in CLIP-ViT-B on SSv2-Small dataset. Skip means adding \emph{D$^2$ST-Adapter} every other stage.}
\resizebox{\linewidth}{!}{
\begin{tabular}{l|c|c|cc}
\toprule
\multirow{2}{*}{Insertion Position} & Tunable & \multirow{2}{*}{Memory Usage} & \multirow{2}{*}{1-shot} & \multirow{2}{*}{5-shot} \\
 & Params (\%) & & & \\
\midrule
Early-insertion & 4.1\% & 15.7 GB & 48.4 & 64.6 \\
Late-insertion  & 4.1\% & 15.3 GB & 54.2 & 68.5 \\
Skip-insertion  & 4.1\% & 15.4 GB & 53.3 & 67.9 \\
Full-insertion  & 7.9\% & 17.3 GB & \textbf{55.0} & \textbf{69.3} \\
\bottomrule
\end{tabular}
}
\label{table_inserted_position}
\end{table}

\begin{table}[t]
\centering
\caption{Effect of the bottleneck ratio of \emph{D$^2$ST-Adapter} in CLIP-ViT-B on SSv2-Small dataset.}
\resizebox{0.9\linewidth}{!}{
\begin{tabular}{l|c|c|cc}
\toprule
\multirow{2}{*}{Ratio} & Tunable & \multirow{2}{*}{Memory Usage} & \multirow{2}{*}{1-shot} & \multirow{2}{*}{5-shot} \\
 & Params (\%) & & & \\
\midrule
0.0625 & 1.4\%  & 15.5 GB & 53.4 & 68.1 \\
0.125  & 3.2\%  & 16.2 GB & 54.2 & 68.6 \\
0.25   & 7.9\%  & 17.3 GB & \textbf{55.0} & 69.3 \\
0.5    & 20.2\% & 19.7 GB & 54.7 & \textbf{69.4} \\
\bottomrule
\end{tabular}
}
\label{table_bottleneck_ratio}
\end{table}

\begin{table*}[t]
\centering
\caption{Performance of different methods using smaller backbones (\emph{i.e.}, ResNet-18 and ResNet-34) on SSv2-Full dataset.}
\resizebox{0.8\linewidth}{!}{
\begin{tabular}{l|ccccc|ccccc}
\toprule
\multirow{2}{*}{Method} & \multicolumn{5}{c|}{{ResNet-18}} & \multicolumn{5}{c}{{ResNet-34}} \\
 & 1-shot & 2-shot & 3-shot & 4-shot & 5-shot & 1-shot & 2-shot & 3-shot & 4-shot & 5-shot   \\
\midrule
OTAM       & 39.4 & 45.0 & 46.6 & 47.4 & 49.0 & 40.6 & 45.2 & 48.0 & 48.9 & 49.2 \\
TRX        & 29.9 & 38.2 & 44.0 & 48.2 & 50.3 & 32.4 & 41.6 & 47.7 & 52.0 & 53.5 \\
HyRSM      & 46.6 & 54.7 & 58.7 & 60.7 & 61.1 & 50.0 & 57.5 & 61.9 & 63.3 & 64.8 \\
MoLo       & \underline{50.0} & \underline{57.2} & 61.6 & \underline{63.6} & 64.6 
           & \underline{54.1} & \underline{61.1} & \underline{65.9} & \underline{67.3} & \underline{67.8} \\
ST-Adapter & 47.3 & 54.3 & 58.1 & 61.3 & 62.2 & 48.9 & 55.6 & 59.4 & 62.5 & 64.1 \\
\midrule
DST-Adapter (Ours)              & \underline{50.0} & 56.9 & \underline{61.7} & 63.4 & \underline{65.3} & 52.1 & 59.3 & 63.4 & 65.8 & 67.5 \\
\textbf{D$^2$ST-Adapter} (Ours) & \textbf{53.0} & \textbf{60.4} & \textbf{65.0} & \textbf{67.1} & \textbf{68.6} 
                                & \textbf{54.4} & \textbf{62.5} & \textbf{66.0} & \textbf{69.1} & \textbf{70.6} \\
\bottomrule
\end{tabular}
}
\label{table_smaller_backbone}
\end{table*}

\begin{table*}[t]
\centering
\caption{Performance of different methods using a larger backbone (CLIP-ViT-L/14). The number of backbone parameters (Params) is also provided as reference.}
\resizebox{0.58\linewidth}{!}{
\begin{tabular}{l|c|cc|cc}
\toprule
\multirow{2}{*}{Method} & \multirow{2}{*}{Backbone (Params)} & \multicolumn{2}{c|}{{SSv2-Full}} & \multicolumn{2}{c}{{Kinetics}} \\
 & & 1-shot & 5-shot & 1-shot & 5-shot \\
\midrule
Full Fine-tuning & CLIP-ViT-L/14 (303.2 M) & 55.5 & 73.0 & 91.7 & 96.8 \\
ST-Adapter       & CLIP-ViT-L/14 (303.2 M) & 67.0 & 83.8 & 92.2 & 96.9 \\
D$^2$ST-Adapter  & CLIP-ViT-L/14 (303.2 M) & \textbf{69.6} & \textbf{85.6} & \textbf{92.8} & \textbf{97.2} \\
\bottomrule
\end{tabular}
}
\label{table_larger_backbone}
\end{table*}

\begin{table*}[t]
\centering
\caption{Hyper-parameter settings of our method with ResNet-50 backbone.}
\resizebox{0.8\linewidth}{!}{
\begin{tabular}{l|l|c|c|c|c|c}
\toprule
Backbone & Hyper-parameter & SSv2-Full & SSv2-Small & Kinetics & HMDB51 & UCF101 \\
\midrule
\multirow{10}{*}{ResNet-50} & base learning rate & 2e-3 & 2e-3 & 2e-3 & 2e-3 & 2e-3 \\
 & warmup start learning rate & 4e-4 & 4e-4 & 4e-4 & 4e-4 & 4e-4 \\
 & training episodes & 120000 & 40000 & 10000 & 8000 & 8000 \\
 & warmup episodes & 12000 & 4000 & 1000 & 800 & 800 \\
 & weight decay & 5e-4 & 5e-4 & 5e-4 & 5e-4 & 5e-4 \\
 & batch size & 8 & 8 & 8 & 8 & 8 \\
 & num. adapters per block & 1 & 1 & 1 & 1 & 1 \\
 & num. input frames & 8 & 8 & 8 & 8 & 8 \\
 & bottleneck ratio & 0.25 & 0.25 & 0.25 & 0.25 & 0.25 \\
 & training crop size & 224 & 224 & 224 & 224 & 224 \\
\midrule
\multirow{10}{*}{CLIP-ViT-B} & base learning rate & 1e-3 & 1e-3 & 5e-4 & 1e-3 & 5e-4 \\
 & warmup start learning rate & 2e-4 & 2e-4 & 1e-4 & 2e-4 & 1e-4 \\
 & training episodes & 120000 & 40000 & 3000 & 3000 & 2400 \\
 & warmup episodes & 12000 & 4000 & 300 & 300 & 240 \\
 & weight decay & 5e-4 & 5e-4 & 5e-4 & 5e-4 & 5e-4 \\
 & batch size & 8 & 8 & 8 & 8 & 8 \\
 & num. adapters per block & 1 & 1 & 1 & 1 & 1 \\
 & num. input frames & 8 & 8 & 8 & 8 & 8 \\
 & bottleneck ratio & 0.25 & 0.25 & 0.25 & 0.25 & 0.25 \\
 & training crop size & 224 & 224 & 224 & 224 & 224 \\
\bottomrule
\end{tabular}
}
\label{table_hyper_parameter}
\end{table*}

\begin{table*}[t]
\centering
\caption{Tuned sampling densities (in the form of $<$T, H, W$>$) for aDSTA-S in the spatial pathway and aDSTA-T in the temporal pathway of our \emph{D$^2$ST-Adapter}. Besides, the sampling densities for aDSTA-Uniform used in Table~\ref{table_sampling_densities} for ablation study is also provided.}
\resizebox{0.7\linewidth}{!}{
\begin{tabular}{c|c|c|c|c}
\toprule
\multirow{2}{*}{Backbone} & \multirow{2}{*}{Feature Map} & Sampling Densities & Sampling Densities & Sampling Densities \\
                          &                              & in aDSTA-S       & in aDSTA-T       & in aDSTA-Uniform \\
\midrule
ResNet-50  & $<$8, 56, 56$>$  & $<$2, 8, 8$>$ & $<$8, 4, 4$>$ & $<$4, 4, 4$>$ \\
ResNet-50  & $<$8, 28, 28$>$  & $<$2, 4, 4$>$ & $<$8, 2, 2$>$ & $<$4, 4, 4$>$ \\
ResNet-50  & $<$8, 14, 14$>$  & $<$2, 4, 4$>$ & $<$8, 2, 2$>$ & $<$4, 4, 4$>$ \\
ResNet-50  & $<$8,  7,  7$>$  & $<$2, 2, 2$>$ & $<$8, 1, 1$>$ & $<$4, 4, 4$>$ \\
\midrule
CLIP-ViT-B & $<$8, 14, 14$>$  & $<$2, 4, 4$>$ & $<$8, 2, 2$>$ & $<$4, 4, 4$>$ \\
\bottomrule
\end{tabular}
}
\label{table_configuration}
\end{table*}

\section{Effect of the Inserted Position of \emph{D\texorpdfstring{$^2$}{}ST-Adapter}}
Theoretically, our \emph{D$^2$ST-Adapter} can be inserted into any position of the backbone flexibly. To investigate the effect of the inserted position of \emph{D$^2$ST-Adapter} on the performance of the model, we conduct experiments with four different ways of inserting \emph{D$^2$ST-Adapters} into the pre-trained CLIP-ViT-B backbone (comprising 12 learning stages) on SSv2-Small dataset: a) early-insertion, which inserts the \emph{D$^2$ST-Adapter} into each of first 6 stages (close to the input), b) late-insertion that inserts the \emph{D$^2$ST-Adapter} into each of last 6 stages (close to the output), c) skip-insertion, which inserts the adapter into the backbone every two stages and d) full-insertion that inserts the adapter into each learning stage.
As shown in Table~\ref{table_inserted_position}, late-insertion, namely inserting the proposed \emph{D$^2$ST-Adapters} into the last 6 stages, yields better performance than early-insertion, which implies that adapter tuning is more effective for task adaptation in deeper layers than in the shallower layers. It is reasonable since deeper layers generally capture the high-level semantic features, which are more relevant to task adaptation. Besides, full-insertion achieves the best performance at the expense of slightly more tunable parameters and memory usage.

\section{Effect of the Bottleneck Ratio of \emph{D\texorpdfstring{$^2$}{}ST-Adapter}}
The tunable parameter size is mainly determined by the bottleneck ratio of \emph{D$^2$ST-Adapter}, defined as the ratio of downsampled channel numbers to the initial size. Thus, we can balance between the model efficiency in terms of tunable parameter size and model effectiveness in terms of classification accuracy by tuning the bottleneck ratio. As shown in Table~\ref{table_bottleneck_ratio}, larger bottleneck ratios typically yield more performance improvement while introducing more tunable parameters, and we set the bottleneck ratio to 0.25 in all the experiments based on the results.

\section{More Visualizations}
\begin{figure*}[t]
  \centering
   \includegraphics[width=\linewidth]{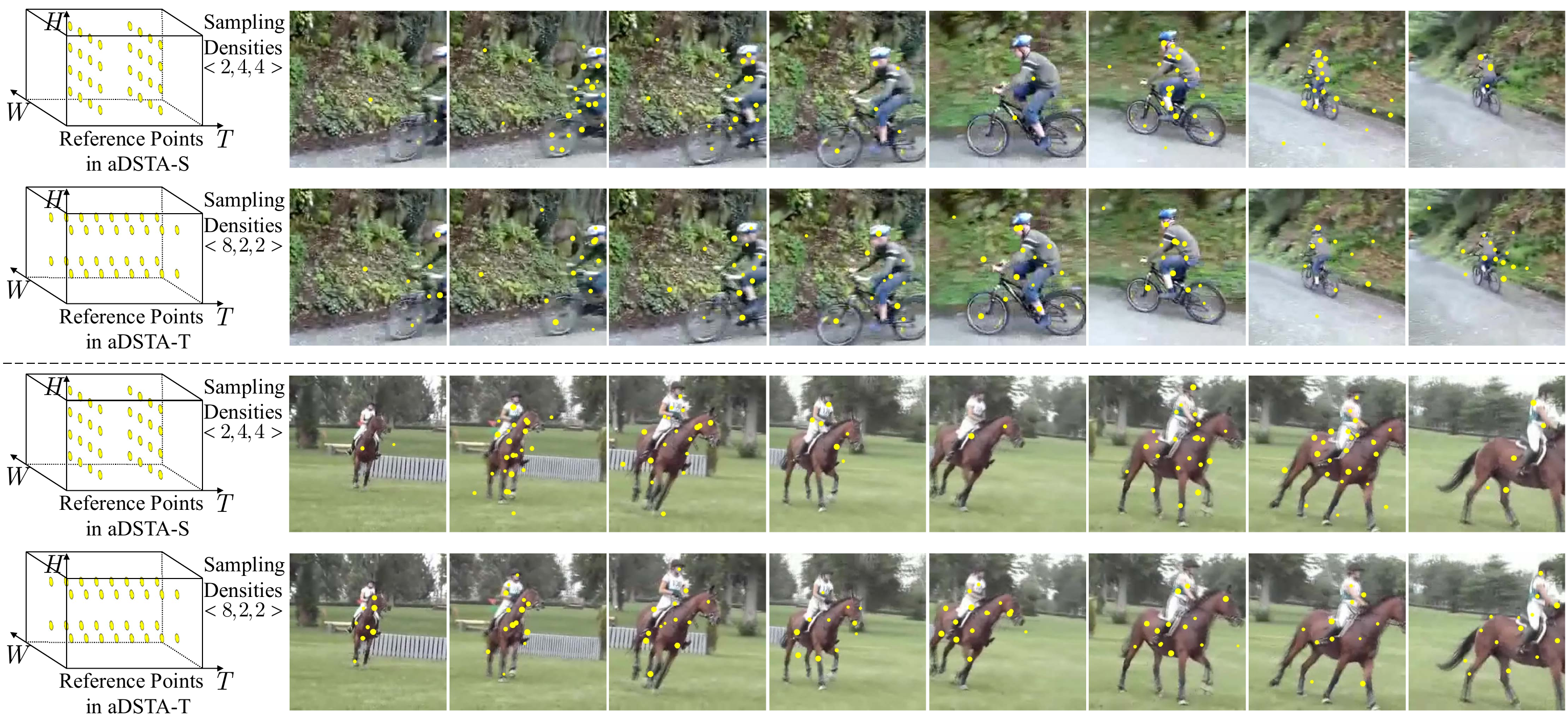}
   \caption{Visualization of top-100 important shifted reference points in both pathways for two video samples. Circle size indicates the importance for each point calculated by aggregating the attention weights from all queries.}
   \label{fig:visualization_supp1}
\end{figure*}

\begin{figure*}[t]
  \centering
   \includegraphics[width=\linewidth]{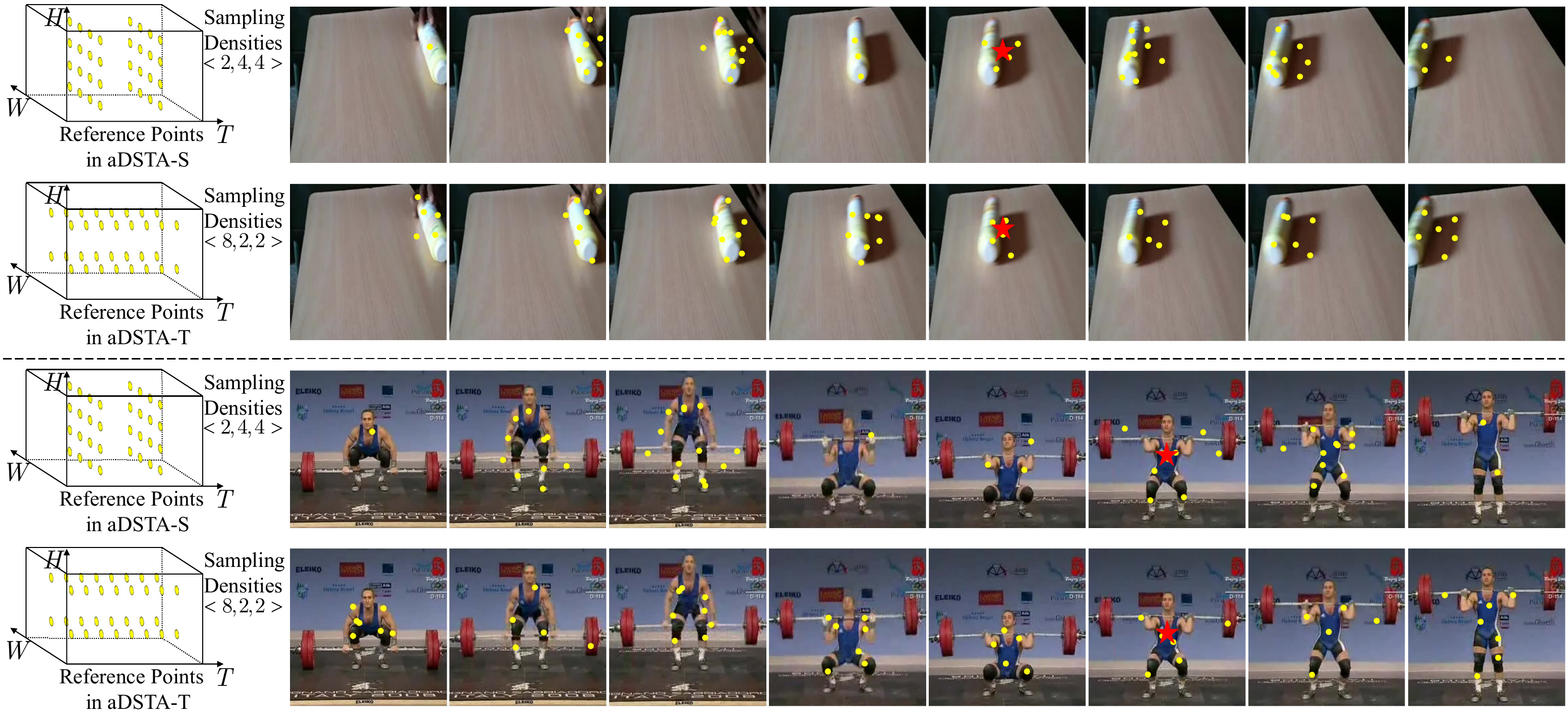}
   \caption{Given a selected query within the salient object indicated by a red star marker, top-50 relevant shifted reference points in terms of attention weight are visualized.}
   \label{fig:visualization_supp2}
\end{figure*}

Consistent with Figure~\ref{fig:visualization} in the paper, we provide more visualization of the shifted reference points in~\Cref{fig:visualization_supp1,fig:visualization_supp2} to evaluate the anisotropic Deformable Spatio-Temporal Attention (aDSTA). Note that only the top-100/50 most important points from the aDSTA modules across all inserted D$^2$ST-Adapters are visualized. The results show that our model is able to capture the salient objects through the shifted reference points in both spatial and temporal domains.

\section{More Implementation Details}
\noindent\textbf{Matching Metric.}
In the few-shot action recognition task, the classification of a query sample is based on the similarities between it and each support class prototype. Typically, a matching metric is first used to temporally align the frames or segments within two videos. Then, the overall similarity between the query and each prototype can be calculated by averaging or summing the similarities of all aligned pairs. We adopt three classic matching metrics in our experiments, including OTAM~\cite{OTAM}, TRX~\cite{TRX}, and Bi-MHM~\cite{HyRSM}. Details of these matching metrics are provided as follows.

OTAM~\cite{OTAM} explicitly leverages the temporal ordering information in videos through tight temporal alignment, which ensures that the alignment results for any two frame pairs do not overlap. It extends the Dynamic Time Warping (DTW) algorithm~\cite{DTW} to compute the alignment path of two videos in the temporal dimension, and employs continuous relaxation to make the model differentiable.

TRX~\cite{TRX} aligns segments rather than individual frames, thus effectively matching actions performed at varying speeds and in different locations across two videos. It first exhaustively enumerates all subsequences of two to four frames as potential actions in both videos. Then, for each action in the query video, the cross-attention mechanism is employed to compute the corresponding action prototype in the support video based on feature similarities, serving as its match.

Bi-MHM~\cite{HyRSM} stands for Bidirectional Mean Hausdorff Metric, which treats the similarity measurement between two videos as a set matching problem. This alignment strategy eliminates the constraints of temporal order and focuses solely on the appearance similarities between frames.

\noindent\textbf{Data Pre-processing.}
We use the same data pre-processing methods as in HyRSM~\cite{HyRSM}. A video is uniformly sampled 8 frames as input. During training, each frame is first resized to 256$\times$256 and then randomly cropped to 224$\times$224. Some basic data augmentation methods are adopted, such as color jitter and horizontal flip. Note that for temporal-related datasets, \ie SSv2-Full and SSv2-Small, we do not use horizontal flip, since the recognition of some classes in these two datasets requires distinguishing between left and right, \eg ``pulling something from left to right''.

\noindent\textbf{Tuned hyper-parameters.}
For ease of the reproduction, we list the tuned values for all the hyper-parameters in our experiments for each dataset with both backbones in Table~\ref{table_hyper_parameter}.

\noindent\textbf{Tuned sampling densities of aDSTA.}
The sampling densities of aDSTA-S in the spatial pathway and aDSTA-T in the temporal pathway of our \emph{D$^2$ST-Adapter} can be tuned on a held-out validation set. Generally, aDSTA-S should sample denser reference points in the spatial domain while aDSTA-T samples more points in the temporal domain. For the instantiation of our model with ResNet, we configure the sampling densities of aDSTA in each convolutional stage individually since the feature maps in different convolutional stages may have different size. In contrast, we only need to tune one configuration for sampling densities when using ViT as the backbone since the feature map always has fixed size in different stages. Table~\ref{table_configuration} shows the tuned configurations of the sampling densities of aDSTA-S and aDSTA-T with different backbones. Besides, we also provide the configurations of sampling densities for aDSTA-Uniform module which is constructed to validate the effect of configuring the sampling densities of aDSTA in Table~\ref{table_sampling_densities}.

\end{document}